\pdfoutput=1
\documentclass{article}

    \usepackage[final]{neurips_2024}

\usepackage[utf8]{inputenc} 
\usepackage[T1]{fontenc}    
\usepackage{amsmath,amssymb}
\usepackage{hyperref}       
\usepackage{cleveref}      
\usepackage{url}            
\usepackage{booktabs}       
\usepackage{amsfonts}       
\usepackage{nicefrac}       
\usepackage{microtype}      
\usepackage{xcolor}         
\usepackage{float}
\usepackage{multirow}
\usepackage{booktabs}
\usepackage{graphicx}

\usepackage{textgreek}
\usepackage{tikz} 
\usepackage{pgfplots} 

\usepackage{amsmath}              
\usepackage{adjustbox}            

\usepackage{times}
\usepackage{epsfig}
\usepackage{graphicx}
\usepackage{amsmath}
\usepackage{amssymb}
\usepackage{fvextra}       

\usepackage{listings}

\definecolor{codegreen}{rgb}{0,0.5,0}
\definecolor{codepurple}{rgb}{0.58,0,0.82}
\definecolor{codegray}{rgb}{0.4,0.4,0.4}
\definecolor{codeblue}{rgb}{0.25,0.5,0.85}

\lstdefinestyle{algpy}{
  language=Python,
  basicstyle=\ttfamily\scriptsize,
  keywordstyle=\color{codepurple}\bfseries,
  commentstyle=\color{codegreen},
  numberstyle=\tiny\color{codegray},
  stringstyle=\color{codeblue},
  numbers=left, numbersep=4pt,
  frame=tb, framerule=0.3pt,
  columns=fullflexible,
  keepspaces=true,
  breaklines=true,
  breakatwhitespace=true,
  showstringspaces=false,
  tabsize=2,
  aboveskip=2pt, belowskip=2pt,
  linewidth=\columnwidth
}

\usepackage[acronym]{glossaries}
\makeglossaries
\newacronym{t2i}{T2I}{text-to-image}
\newacronym{sae}{SAE}{Sparse AutoEncoder}
\newacronym{llm}{LLM}{Large Language Model}
\newacronym{dom}{DOM}{difference-of-means}
\newacronym{nsfw}{NSFW}{Not Safe for Work}

\title{The Unreasonable Effectiveness of Text Embedding Interpolation for Continuous Image Steering}

\author{%
  Yiğit Ekin \\
  Reve\\
  \And
  Yossi Gandelsman \\
  Reve \\
}

\begin{document}

\maketitle
\begin{abstract}
We present a training-free framework for continuous and controllable image editing at test time for text-conditioned generative models. In contrast to prior approaches that rely on additional training or manual user intervention, we find that a simple steering in the text-embedding space is sufficient to produce smooth edit control. Given a target concept (e.g., enhancing photorealism or changing facial expression), we use a large language model to automatically construct a small set of debiased contrastive prompt pairs, from which we compute a steering vector in the generator’s text-encoder space. We then add this vector directly to the input prompt representation to control generation along the desired semantic axis. To obtain a continuous control, we propose an elastic range search procedure that automatically identifies an effective interval of steering magnitudes, avoiding both under-steering (no-edit) and over-steering (changing other attributes). Adding the scaled versions of the same vector within this interval yields smooth and continuous edits. Since our method modifies only textual representations, it naturally generalizes across text-conditioned modalities, including image and video generation. To quantify the steering continuity, we introduce a new evaluation metric that measures the uniformity of semantic change across edit strengths. We compare the continuous editing behavior across methods and find that, despite its simplicity and lightweight design, our approach is comparable to training-based alternatives, outperforming other training-free methods. \footnote{Code is available at \url{https://github.com/YigitEkin/diffusion-sliders}}
\end{abstract}
\section{Introduction}
\label{sec:intro}

\begin{figure}
    \centering
    \includegraphics[width=\linewidth]{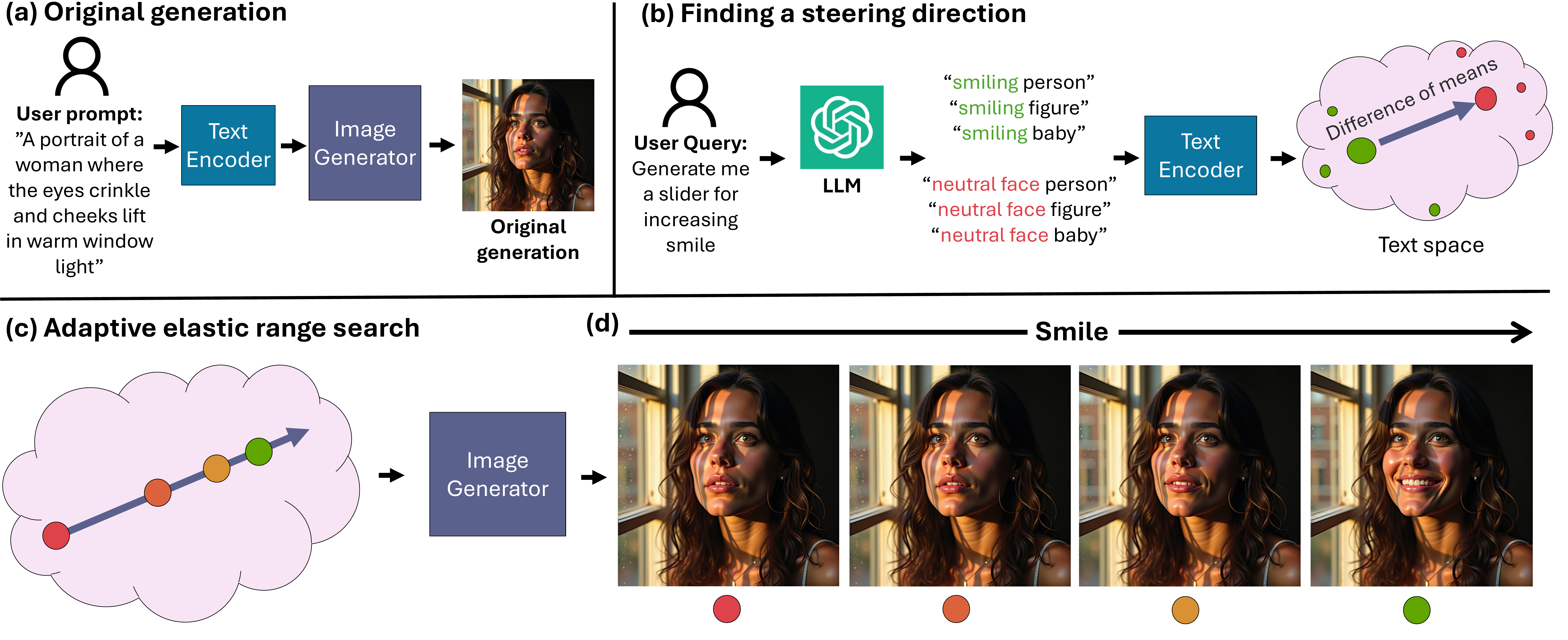}
    \caption{\textbf{Our framework.}
    Given a user text prompt, our method enables controllable editing in text-to-image generation without retraining.
    (a) In the default setting, the prompt is encoded by the text encoder and used by the generative pipeline to produce an image.
    (b) To introduce edit control, we derive a steering direction $\mathbf{d}_s$ by computing a difference-of-means vector from contrastive text pairs (e.g., “smiling person” vs. “neutral face person”).
    (c) We automatically determine the optimal steering strength range via an elastic range search, preventing understeering or oversteering artifacts.
    (d) Such continuous steering along $\mathbf{d}_s$ allows smooth manipulation of the desired attribute (e.g., increasing smile intensity).}
    \label{fig:teaser}
\end{figure}

Text conditioning has become the dominant interface for controlling high-quality visual generation, with strong results across modalities such as image and video synthesis~\cite{dalle2,imagen,ldm,video_diffusion,cogvideox,cogview,opensora,sdxl,wan,lucyedit,flux,flux_kontext,qwen_image}. However, text prompts remain limited when users require \textit{granular} control over semantic attributes (e.g., controlling \textit{how much} a person smiles, or \textit{how photorealistic} an output should be). Such continuous guidance is essential for practical and creative visual workflows, where users often need precise control to obtain the intended result.

A key practical limitation of current continuous editing methods is that they are often not plug-and-play. Many approaches rely on model fine-tuning~\cite{instructpix2pix,concept_slider,flux_kontext,kontinuous_kontext,imagic}, auxiliary trainable modules~\cite{saedit,dravidinterpreting,slideredit}, or architecture-specific designs~\cite{fluxspace}, which makes adaptation to upcoming models increasingly costly in a fast-moving generative ecosystem. Even after selecting a method, using it often requires additional manual setup: beyond specifying the desired edit, users and practitioners typically need to construct contrastive data, define attribute pairs, and choose various method-specific hyperparameters. These methods often are very sensitive to different editing range choices, and an incorrect choice often leads to under or over-editing and an additional trial-and-error refinement process that hurts usability.

In this work, we revisit continuous image editing from a simpler perspective. Our key observation is that, as text-conditioned generative backbones continue to improve, many of the complex procedures and auxiliary trainable modules used in prior work can often be avoided, and  \emph{a linear intervention in the text-encoder representation is often sufficient enough to achieve continuous control in edits}. Motivated by the fact that textual conditioning is the common control interface across text-conditioned generation models, we operate directly on the text-encoder representations. Building on the linear representation hypothesis~\cite{linear_representation} and prior work on steering vectors in language models~\cite{steering_llama,actadd,linear_representation,refusal}, we steer generation by shifting the prompt representation in the text encoder along a concept-specific direction.

We present a text-encoder steering framework (\cref{fig:teaser}) that converts a user-defined concept into a continuous editing slider, without human-in-the-loop intervention. Given an edit concept (e.g., ``smile''), we query a Large Language Model to automatically generate a small set of debiased contrastive prompt pairs (e.g., ``an image of a smiling person'', ``an image of a frowning person''). Using the frozen text encoder of the generative model, we compute a steering direction in text-embedding space by pooling the relevant token representations identified by the Large Language Model and taking their Difference-of-Means~\cite{dom}. This direction is then added to the prompt representation to steer generation along the desired semantic axis.

To enable reliable \textit{continuous} control, we further introduce an elastic range search algorithm to automatically identify an effective set of steering coefficients, avoiding both under-steering (negligible change) and over-steering (degraded or off-manifold outputs). Scaling the same steering vector within this coefficient set yields smooth and continuous edits. We propose a new evaluation metric to evaluate the \emph{uniformity} of semantic change across edit strengths to enable principled comparison of slider behavior across methods. We find that our continuous editing method achieves strong edit compliance while remaining competitive in content preservation and slider continuity, outperforming prior training-free baselines and showing competitive performance with training-based controllers on stronger backbones.

Because our method operates entirely in text-embedding space, it is lightweight and readily applicable to a broad class of text-conditioned generators, including both image and video models. We evaluate our approach against training-based and training-free baselines and show competitive controllability without additional optimization. Our results suggest that simple text-space steering constitutes a strong and practical baseline for future continuous editing methods.

To summarize, our contributions are: 
\begin{itemize}
    \item We propose a framework for continuous and controllable visual editing by steering only the \textbf{text-encoder representations} of text-conditioned generative models. Despite its simplicity, the framework achieves performance comparable to training-based alternatives.
    
    \item We introduce an \textbf{LLM-automated pipeline} that converts a user-defined concept into a usable editing direction by generating debiased contrastive prompt pairs and selecting relevant token representations for steering-vector construction.
    
    \item We propose an \textbf{elastic range search} algorithm that automatically identifies an effective interval of steering magnitudes, enabling smooth continuous control while avoiding under-steering and over-steering.
    
    \item We introduce a new \textbf{continuity evaluation metric} that quantifies the uniformity of semantic change across edit strengths, enabling principled comparison of continuous slider behavior across methods.
\end{itemize}
\section{Related Work}
\label{sec:related_work}

\textbf{Text-conditioned visual generation.}
Text conditioning has become the standard interface for modern visual generation, across both diffusion-based and flow-matching-based models~\cite{dalle2,imagen,ldm,sdxl,video_diffusion,flux,flux_kontext,wan,qwen_image,vace,goku,lucyedit}. 
Flow matching~\cite{lipman2023flow} in particular has emerged as a strong paradigm for high-quality generation, learning continuous velocity fields that map noise to data through an ordinary differential equation (ODE). 
Recent text-conditioned flow-matching models have demonstrated improved sampling efficiency, consistency, and fidelity~\cite{flux,flux_kontext,wan,qwen_image,vace,goku,lucyedit}. 
Our work is complementary to these advances: rather than modifying the image/video generator architecture or training objective, we operate only in the text-conditioning. 
This makes our method compatible with a broad range of text-conditioned backbones, including both diffusion and flow-based models for image and video generation. \\
\textbf{Continuous control in image editing.}
Text-guided editing models such as InstructPix2Pix~\cite{instructpix2pix}, LEDITS++~\cite{ledits++}, Qwen-Image~\cite{qwen_image}, and Flux-Kontext~\cite{flux_kontext} have demonstrated strong capabilities for instruction-based image editing. 
However, controlling the \emph{strength} and \emph{extent} of an edit remains a central challenge. Several works address this by introducing slider-like control through learned modules or latent directions. Flux-Slider~\cite{concept_slider} trains a concept-specific LoRA~\cite{lora} and uses its scale to modulate edit intensity, while SAEdit~\cite{saedit} learns a sparse autoencoder over textual embeddings to find steerable latent directions. Recent instruction-based methods are based on additional training: Kontinuous Kontext~\cite{kontinuous_kontext} adds an explicit edit-strength scalar to an instruction-based editor and trains a lightweight projector to inject this signal into the model.  SliderEdit~\cite{slideredit} provides fine-grained instruction-level control by separating multi-part edit instructions and training lightweight low-rank adapters with a partial prompt suppression objective. In a related direction that also operates directly on textual embeddings via linear semantic directions, Baumann et al.~\cite{baumann2025continuous} identify token-level directions in CLIP space for continuous attribute control, and further train at test-time to correct these directions using diffusion noise predictions across timesteps and Imagic \cite{imagic} does a 2 stage text embedding followed by generator tuning at test time. While these methods provide strong controllability, they introduce additional trainable components or test-time optimization and either remain tied to the target editing architecture or suffer from slower inference time. Closer to our setting, FluxSpace~\cite{fluxspace} achieves training-free control for Flux~\cite{flux} by manipulating architecture-specific internal activations. Our method differs in two key ways: First, we intervene solely in text-encoder representations, without auxiliary trainable modules or test-time correction, which makes the approach easier to transfer across generators. Second, we use an LLM-guided pipeline to automate contrastive prompt-pair construction, token selection, and edit-range calibration, reducing manual work for each new edit concept.
\\ \textbf{Language model steering.}
Steering vectors have been used to control language model behaviors such as sentiment, formality, and refusal tendencies~\cite{steering_llama,actadd,linear_representation,refusal,spectral,angular,entropy}. 
These methods identify linear directions in the representation space of autoregressive transformers and typically apply them to the final token representation to influence the decoding behavior. 
In our setting, however, the text encoder output is provided to the image generator as conditioning context rather than as a next-token prediction state~\cite{t5,qwen3vl,mistral}. As a result, semantic information is distributed across multiple token embeddings instead of being concentrated in a single final-token representation, making last-token steering not directly applicable. 
To adapt steering to this setting, our framework uses an LLM to automatically identify the relevant tokens and injects steering vectors into their embeddings. 
This yields continuous and interpretable control over visual generation without modifying the underlying generative model.
\section{Method}
\label{sec:methodology}
We present a training-free method for continuous steering of text-conditioned visual generation.
Given a user-defined concept (e.g., \emph{happy}$\leftrightarrow$\emph{sad}, \emph{photorealistic}$\leftrightarrow$\emph{cartoon}), we use a Large Language Model to produce a small, debiased set of contrastive prompt pairs and compute a steering vector via token-level pooling and difference-of-means (\cref{sec:method-steer}).
Relevant tokens are automatically selected using an Large Language Model assisted procedure (\cref{sec:token_selection}), and an elastic range search determines effective steering coefficients for smooth, continuous control (\cref{sec:method-continuous}).
Together, these components yield a simple, general framework for controllable generation without training or human supervision.

\subsection{Finding a steering direction}
\label{sec:method-steer}
We start by presenting how language conditioning is used in flow matching models with a text encoder. Then, we explain how to extract steering direction in the representation space of the text encoder. 

\subsubsection{Text-conditioned flow matching.}
We consider a text-conditioned flow-matching generative model that evolves a sample $\mathbf{x}_t$ according to a velocity field $\mathbf{v}_\theta$, conditioned on a text prompt $\mathbf{p}$ encoded by a text encoder $\mathbf{E}$:
\begin{equation}
\dot{\mathbf{x}}_t = \mathbf{v}_\theta(\mathbf{x}_t, t, \mathbf{E(p)}),
\label{eq:flow}
\end{equation}
where $\dot{\mathbf{x}}_t$ denotes the instantaneous change of $\mathbf{x}_t$ at time $t$.
To modulate a target semantic concept $\mathbf{s}$ (e.g., \textit{smiling}, \textit{realism}), we steer the conditioning embedding obtained from the text encoder:
\begin{equation}
\mathbf{E(p)}' = \mathbf{E(p)} + \alpha\,\mathbf{d}_{\mathbf{s}},
\label{eq:steering}
\end{equation}
where $\mathbf{d}_{\mathbf{s}}$ is the learned steering vector and $\alpha$ controls its steering strength.  
Substituting $\mathbf{E(p)}'$ into Eq. \ref{eq:flow} perturbs the generative trajectory along the semantic direction $\mathbf{s}$, enabling fine-grained, text-guided control in the encoder’s output space.

\subsubsection{LLM-generated contrastive pairs.}
To compute steering directions, we construct a set of $K$ contrastive prompt pairs $\{(p_i^+, p_i^-)\}_{i=1}^K$, each differing only in the target concept while preserving all other semantic content. 
For example, for the concept \textit{sad}$\leftrightarrow$\textit{happy}, the corresponding pair may be (\textit{``A portrait of a sad person''}, \textit{``A portrait of a happy person''}). 
These contrastive pairs are automatically generated by a large language model, which ensures that changes are concept-isolated and contextually consistent, eliminating the need for manual dataset design or bias-prone selection. The full query prompt for dataset generation can be seen in the Appendix.

\subsubsection{Debiasing via style-token pooling.}
\begin{figure}[t] 
    \centering
    \includegraphics[width=0.7\linewidth]{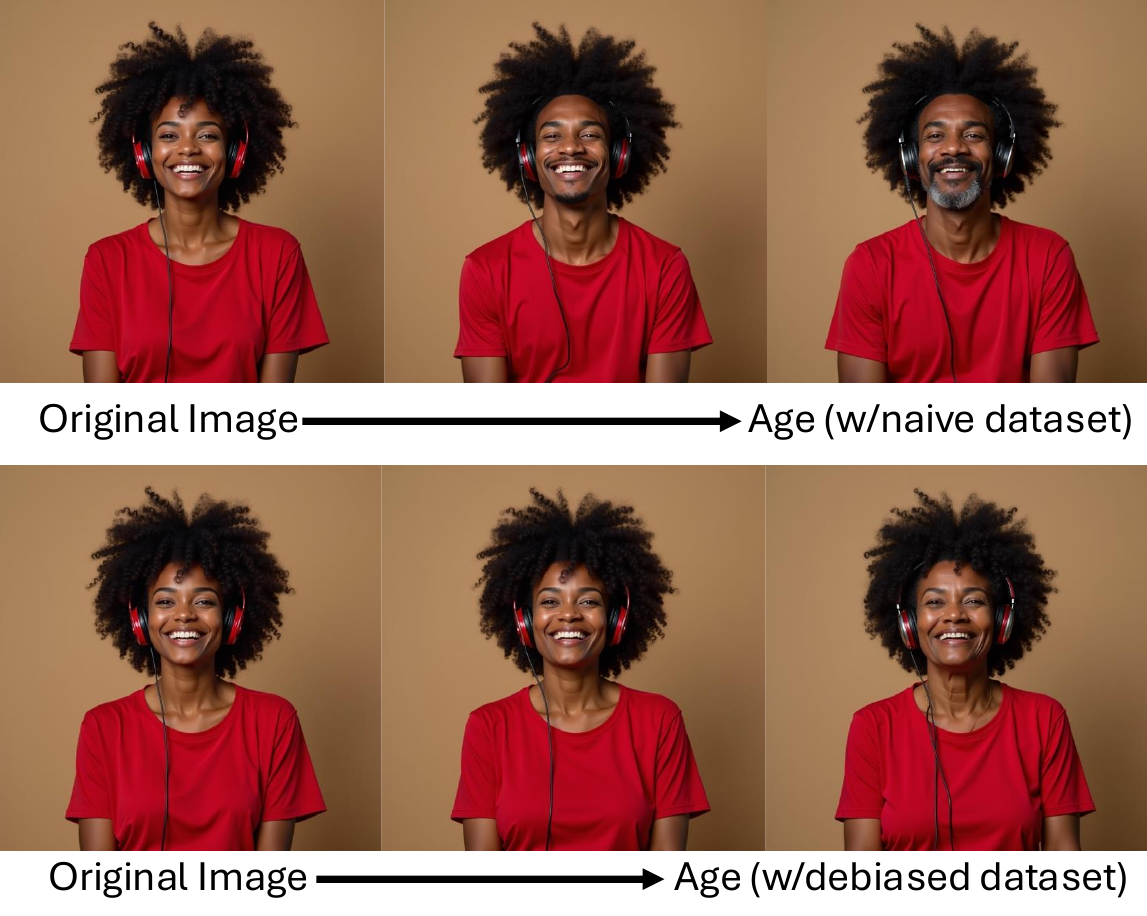}
    \caption{\textbf{Illustration of bias inheritance in steering.} When the age direction is computed from a biased dataset (e.g., predominantly old men), the resulting steering vector entangles gender with age. Consequently, age manipulations not only modify apparent age but also introduce unintended gender-related changes, revealing the dataset’s underlying bias.}
    \label{fig:bias_fig}
\end{figure}

As we rely on an LLM-produced dataset to compute the steering direction $d_s$, a biased dataset can lead to entangled attribute directions. For example, when the contrastive prompts used for constructing the age direction also correlate age with gender (e.g., \textit{"A shot of a young woman"} and \textit{"A shot of a old man"}), the resulting steering vector jointly alters both attributes when manipulating the age direction (top row in \cref{fig:bias_fig}). Such entanglement is undesirable, as the goal is to obtain a steering vector that controls only the intended attribute. To mitigate this, we prompt the LLM to generate balanced contrastive pairs and \textit{only utilize concept-relevant tokens}. E.g., averaging the hidden features of the \textit{young} and \textit{old} token spans in the pair (\textit{``A shot of a young person''} / \textit{``A shot of a old person''}) leads to a more disentangled edit (bottom row in \cref{fig:bias_fig}). 

More precisely, let $S_i^{+}$ and $S_i^{-}$ denote the style-token spans for each pair $i$. The pooled features are
\begin{equation}
{\mathbf{E(p)}}_{i\pm} = \tfrac{1}{|S_i^{\pm}|}\!\!\sum_{j\in S_i^{\pm}}\!\!\mathbf{E(p_i)[j]},
\label{eq:pool_compact}
\end{equation}
which isolates the desired attribute from unrelated tokens (e.g., background or object words), yielding cleaner and more concept-specific directions.

\paragraph{Difference-of-means steering vectors.}
After obtaining the pooled features for each contrastive pair, we compute the difference-of-means vector using the final output representation of the text encoder as
\begin{equation}
\mathbf{s}
= \tfrac{1}{K}\!\sum_{i=1}^{K}\mathbf{E(p)}_{i,+}[j]
- \tfrac{1}{K}\!\sum_{i=1}^{K}\mathbf{E(p)}_{i,-}[j]
\label{eq:dom_compact}
\end{equation}
which captures the semantic displacement between the positive and negative poles of the target concept.  
To ensure consistent magnitude across different concepts and prevent scaling bias during inference, the resulting vector is $\ell_2$-normalized:
$\mathbf{d}_{s} = \frac{\mathbf{s}}{||\mathbf{s}||_2}$.
The normalized direction $\mathbf{d}_{s}$ defines a single, global steering axis applied exclusively to the encoder’s final output embedding.  
Unlike SAEdit~\cite{saedit}, which train sparse autoencoders to discover edit directions, or SliderEdit \cite{slideredit} which learns selective LoRA's in the token space, our method derives a steering vector at test-time directly from the encoder’s final latent space without any need of additional training.

\subsection{Single-image steering}
\label{sec:single_image}
Given an image generated from an input text, our method modifies the text embeddings of selected tokens according to the learned steering vector. The tokens that should be updated are selected by an LLM.

\subsubsection{LLM-assisted token selection.}
\label{sec:token_selection}
We utilize an LLM at test-time to choose which tokens in the free-form input prompt should be steered for a given concept.  
We find that accurate token selection for adding the intended steering vector (e.g., adding \textit{smile} steering vector to the token corresponding to \textit{man}) is crucial for producing semantically aligned edits.  When irrelevant tokens (e.g., stopwords) are modified, the edit may introduce either undesired changes or no changes at all depending on the selection.

To only steer concept-relevant tokens while covering a broad spectrum of edits, we group edits as \emph{local}, \emph{global} or \emph{stylization} and prompts are classified as \emph{implicit} or \emph{explicit}.  \emph{local} edits modify localized attributes of the main subject while preserving subject identity and the overall scene (e.g., \textit{smile}, \textit{age}); \emph{global} edits induce global or geometric changes that affect many regions at once, such as environment or pose (e.g., \textit{winter} $\leftrightarrow$ \textit{summer}, \textit{standing} $\leftrightarrow$ \textit{sitting}); and  \emph{Stylization} edits primarily change the rendering style (e.g., \textit{cartoon}, \textit{anime style}).
\emph{Implicit} prompts are prompts that do not contain the target attribute that is intended to be edited (e.g., ``a man'' for \textit{smile}), while \emph{Explicit} prompts directly include it (e.g., ``a sad man'' for \textit{smile}).

\subsubsection{Rule-based strategy guided by an LLM.}
We automate token selection by prompting an LLM with a simple set of rules: \\
\textit{Stylization edits:} implicit prompts $\rightarrow$ steer main subject nouns (e.g., ``cat''); explicit prompts $\rightarrow$ steer style tokens (e.g., ``cartoon'', ``photorealistic''). \\
\textit{global edits:} implicit prompts $\rightarrow$ steer the main subject nouns (e.g., ``dog'' for \textit{standing}); explicit prompts $\rightarrow$ steer only the global edit tokens (e.g., ``standing'', ``winter'', ``close-up''). \\
\textit{local edits:} implicit prompts $\rightarrow$ steer the main subject nouns (e.g., ``man'' for \textit{smile}); explicit prompts $\rightarrow$ steer only the attribute tokens (e.g., ``sad'', ``old'').

This rule-based LLM-guided procedure removes the need for manual annotation and ensures that steering modifies only semantically meaningful tokens, aligning edit scope with the intended concept. The full query prompt for token selection and inference details for specific edit types can be found in the Appendix.


\subsection{Continuous steering}
\label{sec:method-continuous}

\begin{figure}
    \centering
    \includegraphics[width=\linewidth]{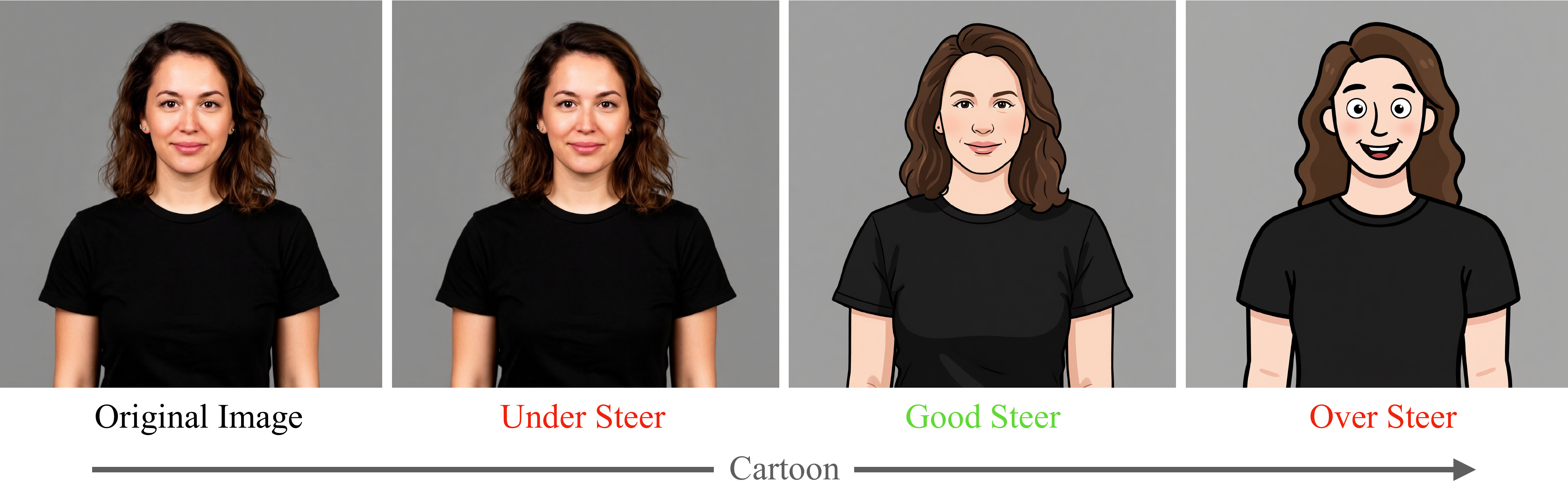}
    \caption{\textbf{Effect of steering magnitude on edit strength.}
Weak steering produces minimal visual change, while large magnitudes lead to semantic drift.
Our elastic range search algorithm automatically identifies the optimal steering range that achieves perceptually consistent edits. In this case, a balanced cartoon stylization.
}
    \label{fig:elastic}
\end{figure}

We aim to create a continuum of image edits by automatically determining the steering strength range that produces smooth and perceptually consistent edits.  
Most existing methods select this range $\alpha \in [\alpha_{\min}, \alpha_{\max}]$ through manual inspection, which often leads to non-continuous results: weak steering produces almost no change, while large magnitudes cause artifacts (see \cref{fig:elastic}).  
Therefore, we present an algorithm that adaptively finds the optimal range for each concept, inspired by elastic band search \cite{elastic_band} (see Listing \ref{lst:elasticband}).

\begin{lstlisting}[style=algpy,
    float=!t,
    caption={\textbf{Elastic Range Search}},
    label={lst:elasticband},
    belowskip=0pt,
    aboveskip=4pt
]
def elastic_band_search(p, g, dist, a_min, a_max, target_gap, eta, lam, eps, 
                        N0, Nmax, Texpand, T, move_fraction, sim_min, sim_max):
    """
     Args:
       p: input prompt; g: image generator where g(p, a) applies steering a;
       dist: perceptual distance function; a_min, a_max: steering range; 
       target_gap: desired perceptual gap; eta: step-size schedule; 
       lam: left/right imbalance gain; eps: minimum spacing; 
       N0, Nmax: initial/max control-point count; Texpand: expansion threshold; 
       T: max iterations; move_fraction: min required movement; 
       sim_min, sim_max: allowed similarity range to the unsteered image.
     Returns: 
       X_valid: control points that satisfy the sim
    """
    X = linspace(0, a_max, N0)
    move_threshold = move_fraction * eps
    for t in range(1, T + 1):
        # Compute normalized perceptual gaps between neighbors
        G = []
        for i in range(len(X) - 1):
            gap = dist(g(p, X[i]), g(p, X[i + 1]))
            G.append(gap / target_gap)
        # EXPAND: insert midpoint where gap is largest
        k = argmax(G)
        if G[k] > Texpand and len(X) < Nmax:
            mid = 0.5 * (X[k] + X[k + 1])
            X = X[:k + 1] + [mid] + X[k + 1:]
            continue
        # MOVE: adjust interior points to equalize gaps
        moved = False
        base_step = eta(t)
        for i in range(1, len(X) - 1):
            L, R = G[i - 1], G[i]
            direction = -1 if L > R else +1
            step = base_step * (1.0 + lam * abs(L - R))
            if direction < 0:
                new_a = max(X[i-1] + eps, X[i] - step)
            else:
                new_a = min(X[i+1] - eps, X[i] + step)
            if abs(new_a - X[i]) >= move_threshold:
                X[i] = new_a
                moved = True
        if not moved:
            break
    # Filter control points by similarity to the unsteered image
    X_valid = []
    for a in X:
        sim = dist(g(p, 0.0), g(p, a))
        if sim_min <= sim <= sim_max:
            X_valid.append(a)
    return X_valid
\end{lstlisting}
\subsubsection{Data-driven initialization.}
The initial steering intervals starting points is estimated directly from the contrastive dataset used to construct the steering vector.  
Given positive pairs in the contrastive dataset $p_i^+$ and the unnormalized steering direction $\mathbf{s}$, we compute empirical projection $\alpha_{\max} = \max_j \langle \mathbf{s}, \mathbf{E(p_i^+)} \rangle$. This limit reflect the actual semantic variation of the concept in the embedding space.  
This data-driven initialization avoids arbitrary heuristic choices and anchors the search to the underlying representation statistics. 
In addition, we have emprically found out that sometimes the text encoder's output space can support further extrapolation beyond the dataset's observed range. To be able to capture this extrapolation if exists, 
we follow a simple extrapolation lookup strategy: we generate an image with double the initial $\alpha_{\max}$ and check if the perceptual distance to the unsteered image is within a reasonable range. If it is, we set $\alpha_{\max}$ to this extrapolated value and continue expanding until the distance is out of the reasonable range or the maximum number of extrapolation steps is reached. This allows us to capture a wider range of edits when the model's representation supports it, while still ensuring that we do not venture into regions of the embedding space that produce semantically inconsistent results.
In practice, over the text encoder's that have been used in modern generative models \cite{t5,qwen3vl,mistral,clip}, we have found out that at most 3 extrapolation steps are needed to capture the possible extrapolation of edits, which adds negligible overhead to the overall search process.

\subsubsection{Elastic-band optimization.} 
The range $[\alpha_{\min}, \alpha_{\max}]$ is split equally into a sequence of control points $\{x_i\}_{i=1}^N$, each corresponding to a generated image at a particular steering strength $\alpha_i$.  
Adjacent points are connected by virtual springs whose tension $f$ encodes the perceptual difference between consecutive generations, measured by an image distance metric (i.e., DreamSim~\cite{dreamsim}).  
The algorithm iteratively alternates between two operations:
\begin{itemize}
    \item \textbf{MOVE:} Each interior point $x_i$ is compared to its neighbors $x_{i-1}$ and $x_{i+1}$.  
    It is moved toward the neighbor associated with the larger normalized distance $f(x_i, x_{i+1})$ or $f(x_{i-1}, x_i)$, redistributing points to equalize perceptual spacing along the curve.  
    The step size follows a cosine-scheduled learning rate that decays over iterations, scaled by the imbalance factor $\lvert f(x_i, x_{i-1}) - f(x_i, x_{i+1}) \rvert$ to accelerate convergence when asymmetry is high.
    \item \textbf{EXPAND:} If a perceptual gap between two adjacent points exceeds a threshold, a midpoint is inserted.  
    This refinement increases sampling density in regions where perceptual change is steep, improving precision without unnecessary evaluations.
\end{itemize}
All shifts are clamped to avoid point order flips.  
Batch rendering enables parallel evaluation of multiple points, substantially reducing runtime.

\subsubsection{Adaptive convergence and output.}
The search proceeds until no further movement or expansion occurs.  
The resulting endpoints $(\alpha_{\min}, \alpha_{\max})$ represent the smallest and largest steering magnitudes that produce perceptually smooth and semantically consistent edits. This adaptive calibration replaces manual tuning with a continuous, data-driven procedure that generalizes across edit categories.
\section{Experiments}
\label{sec:experiments}
We evaluate our method's ability to perform controllable edits across diverse visual concepts. Following previous approaches, we consider three main edit 
categories: \emph{local}, \emph{global}, and \emph{stylization}. Local edits include attributes such as \emph{smile}, \emph{age}, \emph{frostiness},
 \emph{wet texture}, and \emph{rusty surface}. Stylization edits include \emph{cartoon}, \emph{ghibli style}, \emph{anime}, and \emph{photorealistic render 
 style}. Global edits include \emph{winter}, \emph{nighttime}, \emph{rainy weather}, \emph{sitting}, \emph{crowded scene}, and \emph{zoom-out camera} edits. SAEdit~\cite{saedit}, FluxSpace~\cite{fluxspace}, and Flux-Sliders~\cite{concept_slider} are all based on \textsc{Flux.dev}~\cite{flux}; therefore, we report results on this backbone for a consistent comparison. Since Kontinuous Kontext~\cite{kontinuous_kontext} and SliderEdit~\cite{slideredit} are based on the stronger image-to-image \textsc{Flux-Kontext}~\cite{flux_kontext} pipeline, we additionally evaluate our method on a strong instruction-based image editing backbone, \textsc{Qwen-Image-Edit}~\cite{qwen_image}. This experiment is intended to further support our central claim that, as backbones improve, simple linear interpolation in text-embedding space can be sufficient to obtain effective continuous control.  We use DreamSim~\cite{dreamsim} to measure content and identity preservation, and $\Delta$VQA~\cite{vqa} 
to quantify perceptual edit success. To evaluate the continuous behavior of slider edits, we further compare the $\Delta$VQA distance and our proposed 
Monotonic Increment Deviation (MID) metric with 6 number of points. For all experiments, we use a debiased dataset of size 100. See the Appendix for additional details, including 
timestep schedules, steering coefficients, and dataset composition.

\subsection{Monotonic Increment Deviation (MID\textsubscript{dist}) and Gain}
An ideal continuous image editing pipeline should distribute semantic change across the slider in proportion to perceptual change. 
To evaluate this behavior, we generate $N$ edited images at uniformly spaced edit strengths
\begin{equation}
    \alpha_i = \frac{i}{N-1}\alpha_{\max}, \quad i \in \{0,\dots,N-1\},
\end{equation}
where $\alpha_0=0$ denotes the unedited image and $\alpha_{N-1}=\alpha_{\max}$ denotes the strongest edited image.
For each adjacent pair, we compute semantic and perceptual increments:
\begin{equation}
\begin{aligned}
    \Delta v_i &= \left|\Delta\mathrm{VQA}\!\left(I_{\alpha_{i+1}}\right)-\Delta\mathrm{VQA}\!\left(I_{\alpha_i}\right)\right|,\\
    \Delta d_i &= \mathrm{LPIPS}\!\left(I_{\alpha_{i+1}}, I_{\alpha_i}\right),
    \qquad i \in \{0,\dots,N-2\}.
\end{aligned}
\end{equation}
We normalize these increments into distributions over slider steps:
\begin{equation}
    p_i=\frac{\Delta v_i}{\sum_{j=0}^{N-2}\Delta v_j+\varepsilon},\qquad
    q_i=\frac{\Delta d_i}{\sum_{j=0}^{N-2}\Delta d_j+\varepsilon}.
\end{equation}
We then define \textbf{MID\textsubscript{dist}} as the total variation distance between the two distributions:
\begin{equation}
    \mathrm{MID}_{\mathrm{dist}}=\frac{1}{2}\sum_{i=0}^{N-2}\left|p_i-q_i\right|.
\end{equation}
Lower $\mathrm{MID}_{\mathrm{dist}}$ indicates better continuity, i.e., semantic change is distributed more consistently with perceptual change across the slider.

\begin{figure*}[t] 
  \centering
    \includegraphics[width=0.7\textwidth]{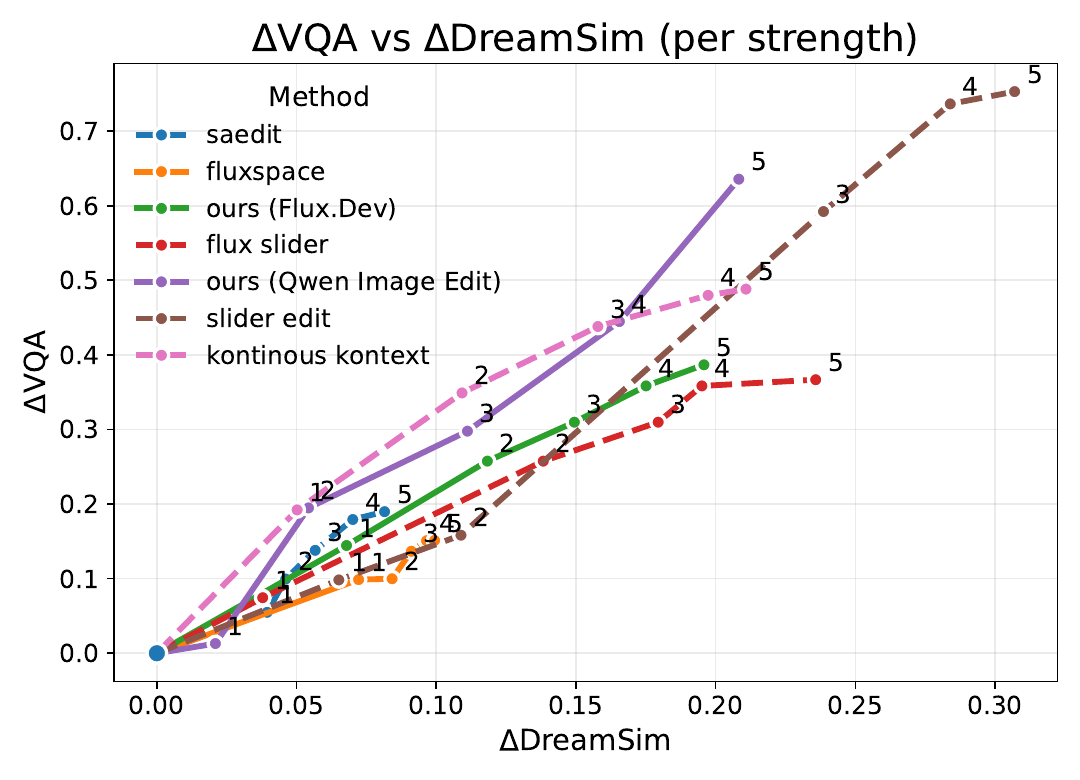}
    \caption{\textbf{Trade-off between edit strength and fidelity.}
    $\Delta$VQA (edit success) vs. DreamSim (distance to the original).
    Curves correspond to different methods; numeric labels denote increasing steering strengths.
    Methods closer to the upper--left corner achieve stronger edits at lower distortion. Training-based methods are given in dashed lines while training-free methods are given in straight lines.}
    \label{fig:sae_plot}
\end{figure*}
\subsection{Qualitative Results}
\label{sec:qualitative}
\begin{figure*}
    \centering
    \includegraphics[width=0.7\linewidth]{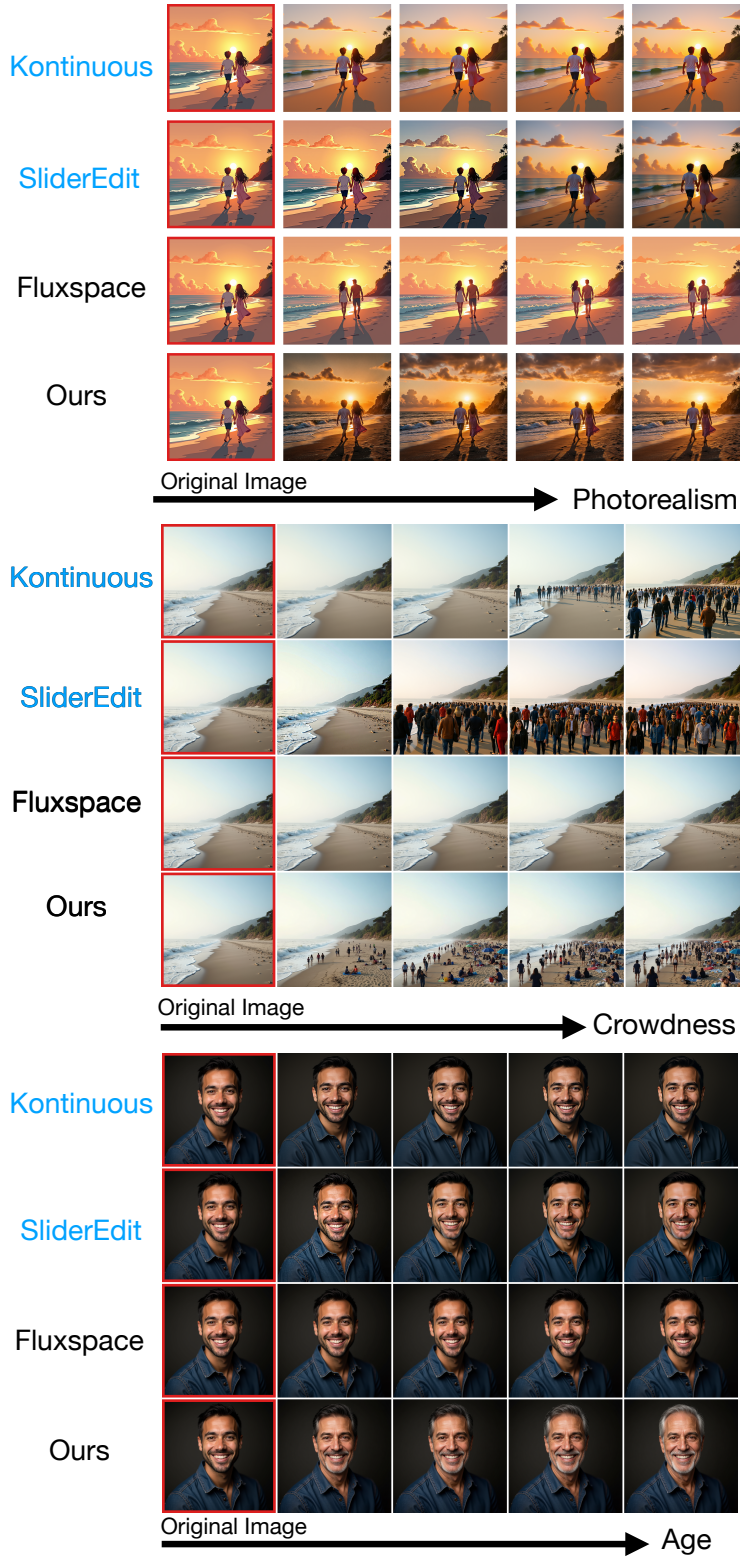}
    \caption{\small \textbf{Qualitative Results.} Our method enables a diverse range of continuous and disentangled semantic edits across various image styles. Leftmost images with red border are initial generations, and from left to right the edit strength is increased. We demonstrate the ability to add style (photorealism), global scene change (crowdness) and local texture change (age). Methods with {\color{cyan}Blue} text are training based methods}
    \label{fig:qualitative}
\end{figure*}

~\cref{fig:qualitative} compares recent training-based methods (Kontinuous Kontext and SliderEdit) and a training-free baseline (FluxSpace) across three edit types: stylization (photorealism), global scene editing (increasing crowdedness), and local editing (increasing age). Across all edits, our elastic-band search selects an initial slider point where the desired effect is already perceptible, which is not consistently observed for the baselines. For photorealism, Kontinuous exhibits an unstable progression with visible artifacts, while SliderEdit shows abrupt, non-smooth changes despite achieving the target effect; FluxSpace under-edits. For age editing, SliderEdit and Kontinuous produce the intended change but with limited strength, whereas FluxSpace largely fails to realize the edit. For crowdedness, Kontinuous and SliderEdit achieve the desired change but do not provide a smooth, slider-like transition, while FluxSpace again fails to produce the edit. Additional qualitative comparisons with other methods are provided in the supplementary material.

\subsection{Quantitative results}
\label{sec:quantitative}

\subsubsection{Methods comparison.}
Tab.~\ref{tab:quant-userstudy} compares training-based and training-free continuous editing methods using two metrics: edit success ($\Delta$VQA$\uparrow$) and slider continuity (MID$\downarrow$). DreamSim and VQA trade-offs across edit strengths are shown in Fig.~\ref{fig:sae_plot}. Per-edit-category statistics are provided in the appendix. Training-based approaches built on the \textsc{Flux.Kontext} backbone achieves the strongest edit compliance, with SliderEdit obtaining the highest $\Delta$VQA. However, this improvement comes with larger perceptual deviation and reduced slider smoothness, reflecting a trade-off between edit strength and controlled, gradual changes. In contrast, architecture- and representation-manipulation methods such as FluxSpace and SAEdit achieve lower DreamSim values and smoother slider behavior (lower MID). As illustrated in Fig.~\ref{fig:sae_plot}, these methods remain closer to the original image across edit strengths, resulting in smaller perceptual distances. However, this behavior largely arises because these approaches often fail to perform stronger global or stylistic edits. While they can handle localized attribute modifications, they struggle with global appearance changes and stylization edits, frequently producing minimal or no visible edits. Consequently, their DreamSim values appear lower in Fig.~\ref{fig:sae_plot}, since the generated images remain closer to the input. Importantly, our training-free steering method benefits from stronger editing backbones. While our method achieves moderate edit success on the Flux.dev backbone, applying the same approach on the stronger Qwen-Image editing backbone increases edit compliance to 0.63 while maintaining competitive preservation and continuity. It can be seen from Fig.~\ref{fig:sae_plot} that our approach lies closest to this region among the compared methods, indicating a favorable balance between edit strength and image preservation. These results further support our claim that as editing backbones improve, lightweight text-space steering becomes an increasingly competitive alternative to specialized training-based control modules, while avoiding additional training or architectural modifications.

\begin{table}
  \centering
  \setlength{\tabcolsep}{2.5pt}
  \renewcommand{\arraystretch}{1.12}
  \resizebox{\linewidth}{!}{%
  \begin{tabular}{l ccccc cccc}
    \toprule
    \textbf{Metric} &
    \multicolumn{4}{c}{\textbf{Training-based}} &
    \multicolumn{5}{c}{\textbf{Training-free}} \\
    \cmidrule(lr){2-5}\cmidrule(lr){6-10}
    &
    \textbf{SAEdit} & \textbf{Flux Slider} & \textbf{SliderEdit} & \textbf{Kontinuous Kontext} & \textbf{FluxSpace} &
    \textbf{Ours (First Token)} & \textbf{Ours (All Tokens)} & \textbf{Ours (Flux.Dev)} & \textbf{Ours (Qwen)} \\
    \midrule
    $\Delta$VQA$\uparrow$  & 0.18   & 0.36 & 0.75   & 0.48   & 0.15   & 0.15   & 0.44 & 0.38      & 0.63 \\
    MID$\downarrow$        & 0.36   & 0.41   & 0.50   & 0.45   & 0.33   & 0.36   & 0.51 & 0.39      & 0.42   \\
    \bottomrule
  \end{tabular}%
  }

  \caption{\textbf{Quantitative and perceptual comparison of editing methods.}
We report $\Delta$VQA$\uparrow$, measuring the increase in edit compliance, and MID$\downarrow$ (Monotonic Increment Deviation), measuring slider monotonicity across consecutive strengths (lower indicates a smoother, more slider-like progression).  Methods are grouped into \textit{training-based} and \textit{training-free} settings.}
  \label{tab:quant-userstudy}
\end{table}

\subsubsection{Runtime Analysis of Proposed Method.}
We report the runtime of each stage of our pipeline in \cref{tab:runtime_breakdown}, averaged over 100 examples per stage. For token selection, we use Qwen3-8B (Thinking) \cite{qwen3}, while GPT-4.1-mini \cite{gpt4} is used for contrastive dataset generation. The elastic search stage is timed with a 30-step generator. It is important to note that dataset generation and token pooling are performed once per edit concept and cached, whereas token selection and elastic search are performed per prompt. During Elastic Band, on average 13.92 images are generated with 3.87 standart deviation. The hyperparameters used in the analysis can be found in appendix.


\begin{table}[t]
    \centering
    \small
    \renewcommand{\arraystretch}{1.15}
    \begin{tabular*}{\linewidth}{@{\extracolsep{\fill}}lccccc@{}}
        \toprule
        \textbf{Stage} & \textbf{Dataset generation (C)} & \textbf{Token pooling (C)} & \textbf{Token selection} & \textbf{Elastic search} \\
        \midrule
        \textbf{Time (s)} & 67.50$\pm$9.77 & 1.09$\pm$0.36 & 3.29$\pm$2.71 & 25.75$\pm$7.85 \\
        \bottomrule
    \end{tabular*}
    \caption{\textbf{Runtime breakdown of our method (seconds)}. It is important to note that Dataset Generation and Token pooling is done per concept and \textbf{cached (C)}, while the token selection and elastic search is done per prompt. On average, a single image generation takes 3 seconds with the used backbones}
    \label{tab:runtime_breakdown}
\end{table}
\subsection{Ablations}
\subsubsection{Effect of token selection.}
To assess the importance of the LLM-based token selection process, we compare our full method, \textbf{Ours (Flux.dev)}, against two alternatives in \cref{tab:quant-userstudy}: \textbf{Ours (First Token)} and \textbf{Ours (All Tokens)}. The results show that token selection is critical for achieving strong edits. When steering is applied only to the first token, the selected representation is weakly correlated with the target concept, so the intervention acts on an uninformative subspace and produces insufficient edits supported by $\Delta$VQA results. In contrast, steering all tokens makes the edit overly global, leading to content drift and degraded preservation. Specifically, steering all tokens results in a DreamSim distance of 0.34, compared to 0.20 when steering the LLM-selected tokens and 0.06 when steering only the first token. Accurate token selection is therefore essential for obtaining sufficient, localized, and content-preserving edits.

\subsubsection{LLM Performance on Token Selection and Dataset Creation.}
We find that token selection does not require a large-capacity model: Qwen3-8B \cite{qwen3} is sufficient to identify the relevant tokens for steering and is therefore used throughout our experiments. In contrast, dataset generation is more sensitive to LLM capability. In our preliminary experiments, smaller LLMs tended to produce more repetitive contrastive examples, reducing diversity in the generated pairs and weakening the debiasing effect of the dataset. For this reason, we use GPT-4.1-mini for contrastive dataset creation. 

\subsubsection{Generalization of Proposed Method to Other Models.}
\label{sec:other_models}
\begin{figure}
    \centering
    \includegraphics[width=\linewidth]{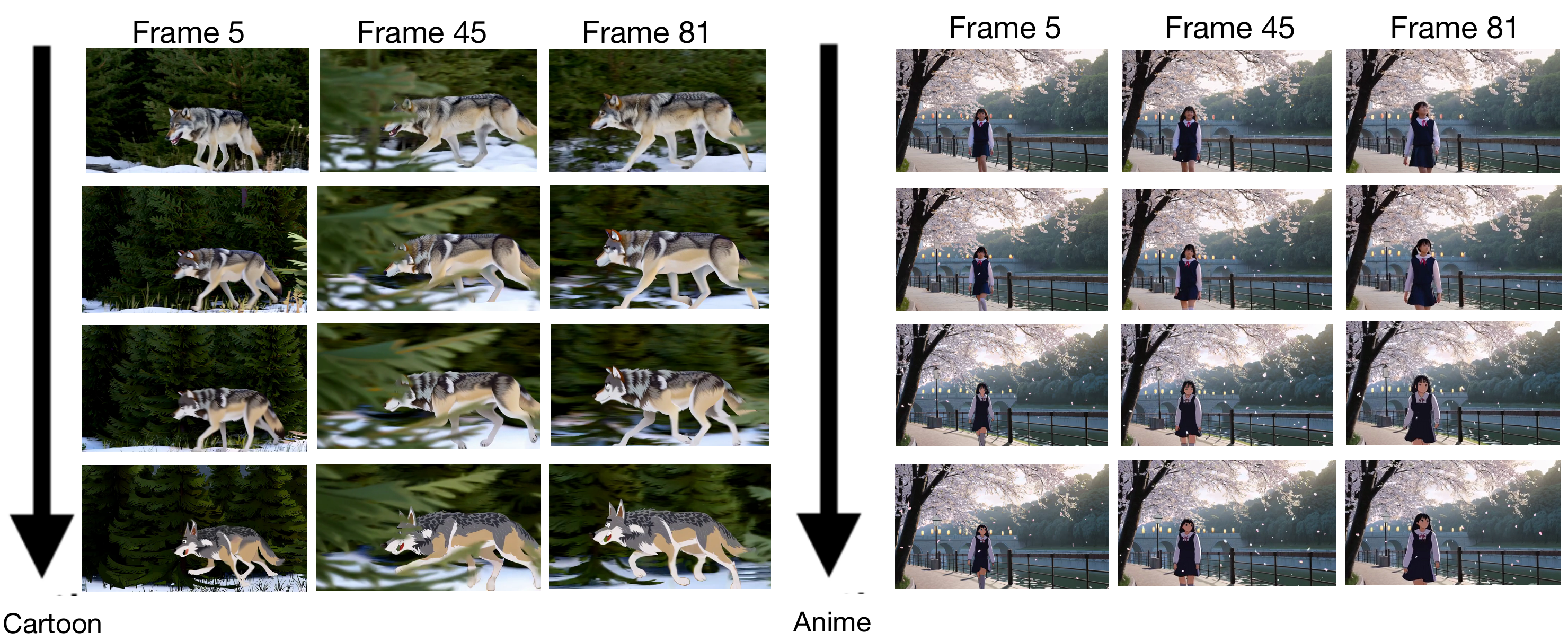}
    \caption{\textbf{Example generations for Wan2.1}  
Because our pipeline operates solely in the text representation, it applies also to video generation models.}
    \label{fig:video}
\end{figure}

Our method operates entirely within the text encoder, making it directly compatible with any diffusion model that relies on
the same text encoder~\cite{t5}. As a result, the same steering vectors can be applied without modification across these models. ~\cref{fig:video} demonstrates this on
a video pipeline~\cite{wan} and on and image-to-image pipeline on \cref{fig:qualitative} showcasing controllable adjustments such as cartoon-style intensity and expression modification. 

\section{Discussion, Limitations \& Future Work}
\label{sec:conclusion}
Our work introduces a training-free framework for controllable image editing through token-aware steering and adaptive range calibration.  
We conclude with limitations of the approach and provide future directions.
\begin{figure}
    \centering
    \includegraphics[width=0.5\linewidth]{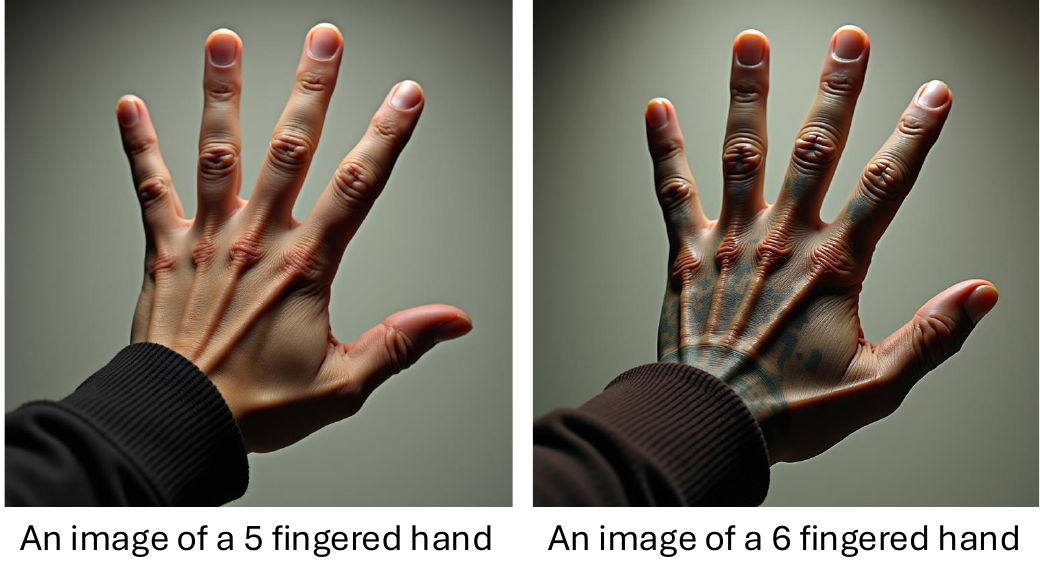}
    \caption{\textbf{Limitations.}Even when prompted with conceptually distinct descriptions (e.g., ``An image of a 5 fingered hand'' vs.\ ``An image of a 6 fingered hand''), the base generative model generates only five-fingered outputs. Such inherent biases prevent us from steering between these two concepts.}
    \label{fig:hand}

\end{figure}
\\
\textbf{Limitations.} Our approach assumes that the generative model can produce both ends of a concept spectrum, allowing interpolation through steering.  
However, this assumption does not always hold in practice due to inherent model biases.  
For example, if the underlying model consistently favors anatomically common structures such as five-fingered hands, it may fail to generate valid six-fingered hands even when explicitly prompted (see \cref{fig:hand}).  
In such cases, the steering direction corresponding to that change cannot be effectively realized.  
Additionally, while token selection can be handled by a relatively small reasoning model, contrastive dataset generation still benefits from a stronger LLM to avoid repetitive examples and maintain diversity in the debiased pairs. As LLMs continue to improve, we expect this dependency to weaken further, which would make our pipeline even more effective and easier to deploy. Finally, according to our knowledge, video models does not have a sufficient enough performing perceptual similarity metric like DreamSim yet. As a result, this prohibits us right now from doing an elastic band search for video generation.

\textbf{Future Work.}
We believe that a more advanced token selection method that dynamically adapts to contextual semantics and cross-token dependencies can improve the results, and we plan to explore this direction. Additionally, developing a faster convergence with better coverage elastic search is a promising future direction. Also, as video editing capabilities of the literature aligns with image-based methods and robust perceptual metrics for video emerge, the integration of this fully autonomous approach to video domain is an exciting future direction. We hope that this work will encourage the community to push controllable generation toward a fully autonomous paradigm, in which interpretable  controls replace fine-tuning and manual dataset curation.
\newpage
\bibliographystyle{plain}
\bibliography{main}
\newpage
\clearpage
\setcounter{page}{1}

\begin{center}
\Large \textbf{The Unreasonable Effectiveness of Text Embedding Interpolation for Continuous Image Steering} \\
\Large Supplementary Material
\end{center}

In addition see \url{https://yigitekin.github.io/diffusion-sliders/supplementary/index.html} for visual results and mp4 video files
\paragraph{Overview.}
This supplementary document provides additional technical details and
extended results that complement the main paper.  Its contents are
organized as follows:
\begin{itemize}
    \item Hyperparameters and implementation details for the elastic-range search.
    \item Details on competing methods for comparison.
    \item Alignment analysis comparing our vectors to those from SAEdit.
    \item Additional qualitative results 
    \item System prompts used for LLM-based token selection and
          contrastive dataset construction. enabled by our method.
    \item Description of the evaluation dataset.
    \item Per Edit Category Distribution of Quantitative Metrics
    \item Inference Details for specific edit types
    \item Hyperparameters Used in Runtime Calculation
    \item Text-to-Image and Image-to-Image Comparison
    \item Dreamsim Scores of Quantitative Comparison
\end{itemize}

\section*{Elastic-Range Search Hyperparameters}

To determine a reliable steering range $[\alpha_{\min}, \alpha_{\max}]$ for each
concept, we use the elastic-range search procedure described in the main paper.
This procedure iteratively samples, expands, and adjusts a set of control points
along the steering axis so that (i) edits progress smoothly, and (ii) the
perceptual deviation from the reference image increases at an interpretable and
consistent rate. For all experiments reported here, these hyperparameters are
used with the \textsc{Flux.dev} backbone.

For transparency, we describe the hyperparameters used by the search algorithm
and explain their role in simple terms.

\paragraph{1. Shared search setup.}
The following hyperparameters are shared across both local edits and
global/stylization edits:
\begin{itemize}
    \item \textbf{height = 512, width = 512:}  
    All images used during elastic-range search are generated and evaluated at
    a resolution of $512 \times 512$.

    \item \textbf{target\_gap = 0.25:}  
    The search aims to keep the perceptual change between neighboring control
    points close to this target value, so that the discovered slider produces
    a reasonably smooth progression of edits.

    \item \textbf{maximum\_number\_of\_iterations = 25:}  
    The maximum number of refinement iterations allowed for the search.

    \item \textbf{expand\_threshold = 0.1:}  
    If the current spacing between neighboring points is too uneven relative to
    the desired progression, the algorithm inserts additional points to better
    resolve that region.

    \item \textbf{maximum\_number\_of\_points = 10:}  
    A hard cap on the total number of control points, preventing the search
    from growing excessively expensive.

    \item \textbf{starting\_number\_of\_points = 4:}  
    The search begins from four initial steering values before refinement.

    \item \textbf{lam = 1.0:}  
    Balances the update objective used during point adjustment.

    \item \textbf{epsilon = 0.01:}  
    A small numerical stabilization constant that prevents degenerate updates.

    \item \textbf{move\_fraction = 0.5:}  
    Controls how aggressively interior points are moved during each update step.
    In practice, this makes the procedure responsive while still stable.

    \item \textbf{minimum\_value = 0.0, maximum\_value = 100.0:}  
    These define the full search interval for the steering coefficient.

    \item \textbf{inference\_batch\_size = 20:}  
    Number of candidate images generated together during each search step.

    \item \textbf{embed\_batch\_size = 64:}  
    Batch size used when computing perceptual embeddings for similarity-based
    evaluation.
\end{itemize}

\paragraph{2. Local edit configuration.}
For local edits, we use:
\begin{itemize}
    \item \textbf{minimum\_similarity = 0.05}
    \item \textbf{max\_similarity = 0.15}
\end{itemize}
The lower minimum similarity for local edits reflects the fact that these edits
should remain relatively close to the original image while still producing a
visible semantic change. In other words, the search accepts milder deviations
from the reference image because local edits are expected to preserve most of
the original content and layout.

\paragraph{3. Global and stylization edit configuration.}
For global edits and stylization edits, we use:
\begin{itemize}
    \item \textbf{minimum\_similarity = 0.15}
    \item \textbf{max\_similarity = 0.30}
\end{itemize}
Compared with local edits, the higher minimum similarity threshold reflects the
fact that global and stylization changes are expected to induce a stronger
perceptual deviation from the original image. These edits typically affect a
larger portion of the image, such as the overall appearance, illumination, or
rendering style, so the valid slider range is searched in a region with more
pronounced change.

\paragraph{Summary.}
Overall, the elastic-range search uses the same optimization and batching
configuration for all edit types in \textsc{Flux.dev}, with the main difference
being the similarity interval used to define a valid range. For local edits, we
use a similarity range of $[0.05, 0.15]$, while for global and stylization
edits we use $[0.15, 0.30]$. This provides a stable and largely
concept-agnostic procedure for automatically identifying usable steering ranges
without manual tuning on a per-concept basis.

\section*{Competing Methods}
We compare the continuous editing capabilities of our method with
SAEdit~\cite{saedit}, Flux-Slider~\cite{concept_slider}, FluxSpace~\cite{fluxspace},
SliderEdit, Kontinuous Kontext, and our image-editing variant based on Qwen Image Edit.
Below, we explain the setup used for each of these methods along with implementation
details where necessary.

Unless otherwise noted, SAEdit, Flux-Slider, and FluxSpace are evaluated using the
standard evaluation prompt set described in \ref{supp:dataset}. In contrast,
SliderEdit, Kontinuous Kontext, and Ours (Qwen Image Edit) are evaluated using an
instruction-based prompt format, since these are image-editing methods rather than
pure text-to-image generation methods. For these instruction-based methods, we first
generate the source images from the prompts in the dataset using \textsc{Flux.dev},
and then use these generated images as the image-conditioning input for the editing
model. The editing instructions are written in the form: \emph{``make the scene
$\langle$edit$\rangle$ without changing the overall format''} or \emph{``without changing the background and the identity of the [person,$\langle$object\_of\_interest$\rangle$ make the [person,$\langle$object\_of\_interest$\rangle$,
$\langle$edit$\rangle$''} depending on the edit and the object of interest.

\paragraph{SAEdit}
For SAEdit, we use their released code implementation. As there is no clear
guideline released by the authors on how to construct a dataset for calculating
steering vectors, we feed their released example datasets to an LLM and ask it
to generate the remaining concepts in a similar format, following the dataset
structure described in \ref{supp:dataset}. For token steering, we steered the same token selections as our method.

\paragraph{FluxSlider}
Although Concept Sliders was originally developed for UNet-based diffusion
models~\cite{concept_slider,sdxl}, we use the extension of this method to
rectified-flow transformers~\cite{flux} for a fairer comparison. Specifically,
we use the Flux slider implementation released by the authors\footnote{\url{https://github.com/rohitgandikota/sliders/tree/main/flux-sliders}}.
For training, we follow the default hyperparameters provided in the official
implementation:
\begin{itemize}
    \item Learning rate: 0.002
    \item Learning rate warmup steps: 200
    \item Training steps: 1000
    \item Batch size: 1
    \item LoRA rank: 16
    \item LoRA alpha: 1
\end{itemize}

During quantitative evaluation, we use a maximum slider scale of 5. For dataset
generation when training sliders, we use the system prompt provided in their
official codebase\footnote{\url{https://github.com/rohitgandikota/sliders/blob/main/GPT_prompt_helper.ipynb}}.

\paragraph{FluxSpace}
FluxSpace is already a training-free method~\cite{fluxspace}. Therefore, we
directly use their released implementation for quantitative evaluation without
modification. For both local and stylization edits, we use the default
hyperparameters from their public code\footnote{\url{https://github.com/gemlab-vt/FluxSpace/blob/main/demo.ipynb}}.

For \textbf{stylization-based edits}, the default settings are:
\begin{itemize}
    \item edit start timestep = 0
    \item edit stop timestep = 3
    \item edit global scale = 0.8
    \item edit content scale = 4.0
    \item attention threshold = 0
\end{itemize}

For \textbf{local edits}, the default settings are:
\begin{itemize}
    \item edit start timestep = 3
    \item edit stop timestep = 30
    \item edit global scale = 0.5
    \item edit content scale = 5.0
    \item attention threshold = 0.5
\end{itemize}

During quantitative evaluation, we use a content scale of 5.

\paragraph{SliderEdit}
SliderEdit is an image-editing-based method, so unlike text-to-image methods, it
first applies the edit and then suppresses it to recover a controllable range.
Accordingly, for comparison we evaluate SliderEdit using the scales
[-1, -0.6, -0.2, 0.2, 0.6, 1]. Here, the negative direction corresponds to
suppressing the applied edit, and we use the scale $-1$ when reporting the
maximum $\Delta$VQA result. As with the other image-editing baselines, we use
the same instruction-based prompts in the form described, and the input
image is the \textsc{Flux.dev}-generated image corresponding to the original
dataset prompt.

\paragraph{Kontinuous Kontext}
Kontinuous Kontext is also evaluated in the image-editing setting using the same
instruction-based prompts as SliderEdit and Ours (Qwen Image Edit). We first
generate the source images from the dataset prompts using \textsc{Flux.dev} and
provide these as image-conditioned inputs. For this method, the slider values are
sampled uniformly as \texttt{np.linspace(0, 1, num=n\_edit\_steps)}. During
quantitative evaluation, we use the value 1 as the maximum edit strength.

\paragraph{Ours (Qwen Image Edit)}
For our Qwen Image Edit variant, we use the same instruction-based prompt format
as the other image-editing methods. As with SliderEdit
and Kontinuous Kontext, we first generate the original images from the dataset
prompts using \textsc{Flux.dev} and then use these images as the image
conditioning input. This ensures that SliderEdit, Kontinuous Kontext, and Ours
(Qwen Image Edit) are compared under the same editing setup, while SAEdit,
Flux-Slider, and FluxSpace are evaluated on the standard prompt-based benchmark
described in evaluation dataset description\ref{supp:dataset}.

\section*{Alignment analysis with SAEdit vectors}
SAEdit~\cite{saedit} extracts steering directions using a learned sparse
autoencoder, while our method derives steering vectors in a fully
training-free manner.
To examine how these approaches relate, we compute the cosine similarity between
SAEdit’s vectors and ours for two representative edits (\textit{smile},
\textit{anime}).
As shown in \cref{tab:sae_alignment}, the similarities are close to zero.
This indicates that the two methods discover \emph{different} directions for the
same concept, highlighting that there are multiple valid ways to construct an
effective steering vector, even when the numerical directions do not coincide.

\begin{table}[]
    \centering
    \begin{tabular}{c|c}
         & Cosine Similarity \\
         \hline
        Smile &   0.03\\
         \hline
        Anime & -0.01\\
         \hline
    \end{tabular}
\caption{\textbf{Alignment with SAEdit steering vectors.}  
Cosine similarity between our training-free steering vectors and those produced
by SAEdit’s learned sparse autoencoder for two representative edits.  
Higher similarity indicates that both methods capture comparable semantic
directions despite fundamentally different construction approaches.}
    \label{tab:sae_alignment}
\end{table}

\section*{Additional qualitative results}
Additional qualitative results are given in \url{https://yigitekin.github.io/diffusion-sliders/supplementary/index.html}

\section*{System prompts for token selection and dataset construction}
We rely on two system prompts: (1) one for constructing contrastive
text pairs used to estimate steering vectors, and (2) one for
selecting which tokens in a free-form prompt should be steered at
inference time. The full system prompts can be seen in ~\cref{subsec:dataset,subsec:token_steer}
\section*{Evaluation Dataset Description}
\label{supp:dataset}

\subsection{Prompt for contrastive dataset generation}
\label{subsec:dataset}
\begin{verbatim}
STEER_CONCEPT = "concept here"
NUMBER_OF_EXAMPLES = 100

prompt = f"""
You are an advanced data generation assistant.

Your task is to create a contrastive dataset of 
{NUMBER_OF_EXAMPLES} examples for 
computing a steering vector.

The steering concept to focus on is: {STEER_CONCEPT}

Output exactly {NUMBER_OF_EXAMPLES} JSON objects, one
per line (JSON Lines), with 
no list brackets, no extra commentary, and no markdown.
Each line must be:
{{"pos_style": "<positive identifier>", "neg_style": 
"<negative identifier>", 
"pos": "<positive full sentence>", "neg": "<negative 
full sentence>"}}

---
### 1. Internal Analysis (YOUR FIRST STEP)

Before generating, analyze {STEER_CONCEPT}:
* Is it an Abstract Style? (e.g., "photorealistic vs 
cartoon", 
"bright vs dark", "metal vs wood"). These can apply 
to any subject.
* Is it a Subject-Specific Attribute? (e.g., "smiling
vs neutral" [faces], 
"ripe vs unripe" 
[fruit], "young vs old" [living beings/objects]). These 
are tied to a class of subjects.

Based on your analysis, you MUST follow the correct 
rules from Section 2.

---
### 2. Generation Guidelines (STRICT)

A. Universal Rules (Apply to ALL concepts):
* PARALLELISM: The two sentences in a pair MUST share 
the same syntactic 
skeleton and content words (subject, setting, 
composition, perspective).
* MINIMAL DELTA: The ONLY differences between 
"pos" and "neg" 
are the minimal tokens 
that express the concept contrast (e.g., 
"smiling" <-> "neutral").
* STYLE NEUTRALITY: Do NOT change rendering domain, 
lighting, camera, or layout.
* IDENTIFIERS: Use the SAME "pos_style" and "neg_style" 
identifiers for ALL 

{NUMBER_OF_EXAMPLES} lines. 
These identifiers MUST also appear in the 
corresponding sentences.

B. Content & Subject Rules (CHOOSE A or B based on 
your Analysis):

[RULE SET A] For ABSTRACT STYLES (e.g., cartoon, bright):
* SUBJECT: You MUST vary subjects and settings widely (e.g., 
objects, landscapes, animals, architecture, indoor/outdoor).
* GOAL: Decouple the style from any one context.
* SAFETY: Avoid specific identities (age, gender, race) 
if people/animals are used. Use neutral terms ("person", 
"animal").

[RULE SET B] For SUBJECT-SPECIFIC ATTRIBUTES (e.g.,
smiling, ripe):
* SUBJECT: You MUST focus *only* on the relevant 
subject class 
(e.g., "person" or "face" for smiling; "fruit" or 
"plant" for ripe). 
Do NOT use inanimate objects like statues for concepts like 
"smiling".
* GOAL: Isolate the attribute's effect on its specific 
subject.
* SAFETY: To avoid bias *within* the subject class, 
use neutral, general terms 
(e.g., "person," "face," "figure," "human"). Do NOT 
specify age, gender, race, 
or ethnicity unless it is the *target concept itself*.

---
### 3. Quality Checks (must pass):
* Exactly {NUMBER_OF_EXAMPLES} lines; each is 
valid JSON.
* "pos" and "neg" are grammatical, depictable, 
and differ ONLY by the 
minimal concept tokens.
* The rules from Section 2 (A and B) have 
been correctly followed.

---
### 4. Examples

* Example for "bright vs dark" (Abstract Style - Rule A):
{{"pos_style": "bright", "neg_style": "dark", 
"pos": "A bright living room with large windows.", 
"neg": "A dark living room with large windows."}}
* Example for "smiling vs neutral" 
(Subject-Specific - Rule B):
{{"pos_style": "smiling", "neg_style": "neutral", 
"pos": "A photorealistic portrait of a person with 
a smiling expression.", "neg": 
"A photorealistic portrait of a person with a ,
neutral expression."}}

Now generate {NUMBER_OF_EXAMPLES} JSON 
objects that represent the 
{STEER_CONCEPT} contrast following these 
instructions, one per line.
"""
\end{verbatim}

\section*{Prompt for token selection at inference}
\label{subsec:token_steer}

\begin{verbatim}
You are an expert token selection assistant. Your job is to 
identify the exact tokens from a 
PROMPT that should be steered, 
based on a steering CONCEPT.

DEFINITIONS

CONCEPT TYPE:
	•	Local Edit: A trait that applies to a specific subject 
    
    (e.g., "smile", "age", "ripe", "sad").
	•	Stylization Edit: A rendering style (e.g., 
    "photorealistic", "cartoon", "anime", "dark").
	•	Global Edit: A change to the entire scene’s context or 
    environment (e.g., "winter", "summer", "crowded").

PROMPT TYPE:
	•	Implicit: The prompt is neutral and does not 
    contain words related to the 
    concept 
    (e.g., PROMPT: "a man", CONCEPT: "smile").
	•	Explicit: The prompt contains a word related
    to the concept, usually a 
    positive 
    or negative pole (e.g., PROMPT: "a sad man", CONCEPT: 
    "smile";  
    PROMPT: "a photorealistic man", CONCEPT: "cartoon").

---

RULES (Apply in order)
	1.	If the CONCEPT is a "Global Edit":
	•	-> Output all meaningful content words from the 
    prompt that describe the 
    scene or objects (ignore filler words and punctuation).
	•	Do not include articles, prepositions, or
    conjunctions unless they are 
    semantically part of a named object or phrase.
	2.	If the CONCEPT is a "Stylization Edit":
	•	If the PROMPT is Explicit:
-> Output only the explicit style or appearance words (e.g., 
"photorealistic", "cinematic", "cartoon").
	•	If the PROMPT is Implicit:
-> Output only the main subject nouns that the style can 
logically apply to (e.g., "man", "lighthouse", "forest").
	3.	If the CONCEPT is a "Local Edit":
	•	If the PROMPT is Explicit:
-> Output only the explicit descriptive or emotional
words expressing the 
local attribute (e.g., "sad", "old", "angry", "smiling").
	•	If the PROMPT is Implicit:
-> Output only the subject nouns that the local attribute 
can naturally attach to (e.g., "man", "woman", 
"child", "face", "fruit").
Prefer the main human, animal, or animate entity; 
if none, choose the most 
central object noun in the description.
	5.	General constraints:
	•	Exclude punctuation and purely functional 
    words (articles, prepositions, etc.).
	•	Return only the minimal set of tokens 
    required to attach the concept.
---

EXAMPLES

Global Edit:
	•	PROMPT: "a woman in a park", 
    CONCEPT: "winter" (Global Edit)
	•	OUTPUT: woman park

Stylization Edit:
	•	PROMPT: "a photorealistic lighthouse on a 
    cliff", CONCEPT: "cartoon" 
    (Stylization Edit, Explicit)
	•	OUTPUT: photorealistic
	•	PROMPT: "a lighthouse on a cliff", 
    CONCEPT: "cartoon" (Stylization Edit, Implicit)
	•	OUTPUT: lighthouse

Local Edit:
	•	PROMPT: "a portrait of a sad man", 
    CONCEPT: "smile" (Local Edit, Explicit)
	•	OUTPUT: sad
	•	PROMPT: "a portrait of a man", 
    CONCEPT: "smile" (Local Edit, Implicit)
	•	OUTPUT: man
	•	PROMPT: "a ripe tomato on the vine", 
    CONCEPT: "age" (Local Edit, Explicit)
	•	OUTPUT: ripe
	•	PROMPT: "a tomato on the vine", 
    CONCEPT: "age" (Local Edit, Implicit)
	•	OUTPUT: tomato

---

YOUR TASK

Analyze the following PROMPT and CONCEPT using 
this logic. Provide 
ONLY the specific 
tokens to steer, separated by a single space. 
Do not add any 
commentary, explanation, 
or punctuation.

PROMPT: "an image of a man"
CONCEPT: "smile"
OUTPUT:
\end{verbatim}

\section*{Evaluation Dataset}

We construct our evaluation prompt set to cover a broad range of edit behaviors
while keeping the underlying scene content diverse. The dataset used in our
experiments contains a total of \textbf{198 prompts}. These prompts are grouped
into three high-level categories: \emph{local edits}, \emph{global edits}, and
\emph{stylization edits}. 

The local edit category contains edits that modify a restricted semantic
attribute while preserving the overall scene identity and composition. The
global edit category contains edits that affect scene-level properties or
broader image characteristics, often requiring coordinated changes across
multiple regions of the image. The stylization category contains edits that
primarily change the rendering style or visual domain of the image while
preserving its semantic content. This structure allows us to evaluate whether a
single continuous steering framework can generalize across qualitatively
different types of edits.

\paragraph{Local edits.}
Local edits are designed to change a specific attribute of the subject or object
while leaving most of the image unchanged. In our dataset, this category
includes the edit types \emph{smile}, \emph{age}, \emph{rust level}, and
\emph{wetness}. These edits test whether the method can produce targeted
changes without unnecessarily altering the surrounding scene.

Two example prompts for each local edit type are given below:
\begin{itemize}
    \item \textbf{Smile:}
    \begin{itemize}
        \item ``A professional passport-style headshot of a man wearing a navy suit against a neutral studio background, closed lips and a straight mouth, looking directly at the camera.''
        \item ``A woman seated at a wooden dining table inside a softly lit apartment, looking at the camera with a neutral expression, lips closed, relaxed face.''
    \end{itemize}

    \item \textbf{Age:}
    \begin{itemize}
        \item ``A portrait of a woman standing beside a bookshelf in a cozy living room.''
        \item ``A man standing outdoors in front of a mountain landscape.''
    \end{itemize}

    \item \textbf{Rust level:}
    \begin{itemize}
        \item ``A vintage car parked on a quiet countryside road.''
        \item ``A metal gate standing at the entrance of an apartment.''
    \end{itemize}

    \item \textbf{Wetness:}
    \begin{itemize}
        \item ``A paved road stretching through a rural landscape.''
        \item ``A wooden dock extending into a calm lake.''
    \end{itemize}
\end{itemize}

\paragraph{Global edits.}
Global edits affect broader scene-level structure or image-wide properties.
These edits typically require coherent transformations across multiple parts of
the image rather than a single localized modification. In our dataset, this
category includes \emph{season}, \emph{time of day}, \emph{rain intensity},
\emph{crowd density}, \emph{camera zoom}, \emph{pose sitting/standing}, and
\emph{frostiness}.

Two example prompts for each global edit type are given below:
\begin{itemize}
    \item \textbf{Season:}
    \begin{itemize}
        \item ``A small countryside house surrounded by trees and a grassy yard.''
        \item ``A mountain landscape with a lake and pine trees in the distance.''
    \end{itemize}

    \item \textbf{Time of day:}
    \begin{itemize}
        \item ``A busy downtown street with tall office buildings and traffic lights.''
        \item ``A coastal harbor with fishing boats docked near wooden piers.''
    \end{itemize}

    \item \textbf{Rain intensity:}
    \begin{itemize}
        \item ``A person walking along a city sidewalk carrying a backpack.''
        \item ``A parked car along a residential street with trees and houses.''
    \end{itemize}

    \item \textbf{Crowd density:}
    \begin{itemize}
        \item ``A city square surrounded by shops and cafes with wide walking space.''
        \item ``A train station platform with benches and signage.''
    \end{itemize}

    \item \textbf{Camera zoom:}
    \begin{itemize}
        \item ``A portrait of a woman standing in a large open field with mountains behind her.''
        \item ``A man standing on a rooftop overlooking a city skyline.''
    \end{itemize}

    \item \textbf{Pose sitting/standing:}
    \begin{itemize}
        \item ``A woman inside a living room near a wooden chair and a small side table.''
        \item ``A standing man in a park near a bench under a large tree.''
    \end{itemize}

    \item \textbf{Frostiness:}
    \begin{itemize}
        \item ``A wooden cabin surrounded by pine trees.''
        \item ``A metal railing along a city bridge.''
    \end{itemize}
\end{itemize}

\paragraph{Stylization edits.}
Stylization edits change the visual appearance or artistic rendering of the
image while preserving its semantic structure. In our dataset, this category
includes \emph{cartoon}, \emph{anime}, \emph{Ghibli}, and
\emph{photorealism}. These edits are used to test whether the method can
continuously control appearance without destroying scene identity.

Two example prompts for each stylization edit type are given below:
\begin{itemize}
    \item \textbf{Cartoon:}
    \begin{itemize}
        \item ``A young woman sitting at a small outdoor cafe table in Paris, wearing a red scarf and holding a cup of coffee.''
        \item ``A firefighter standing confidently in front of a red fire truck.''
    \end{itemize}

    \item \textbf{Anime:}
    \begin{itemize}
        \item ``An image of a teenage girl standing in a pool area.''
        \item ``A boy walking through a quiet suburban street at dusk.''
    \end{itemize}

    \item \textbf{Ghibli:}
    \begin{itemize}
        \item ``A scene of a coffee shop surrounded by people drinking coffee and having fun.''
        \item ``A young girl walking along a countryside path with rolling hills.''
    \end{itemize}

    \item \textbf{Photorealism:}
    \begin{itemize}
        \item ``A woman standing beside a vintage car parked on a quiet street.''
        \item ``A cartoon style image of a surfer emerging from the ocean carrying a surfboard.''
    \end{itemize}
\end{itemize}

Overall, this dataset covers a wide range of edit behaviors, from restricted
attribute manipulation to scene-wide transformations and appearance transfer.
Such diversity is important for showing that the proposed method is not limited
to one narrow class of edits, but instead supports continuous steering across
multiple qualitatively different editing regimes.
\section*{Inference Details for Specific Edit Types}

Our inference strategy differs depending on both the model type and the edit type. For text-to-image backbones, there is no explicit image-conditioning mechanism that preserves the structure of a reference generation. Therefore, for \emph{local edits} we apply the steering progressively across denoising timesteps using a linear schedule, so that the early steps primarily determine the coarse scene structure while later steps incorporate the steered local modification. This helps preserve the original layout and low-frequency content while allowing the edit to appear in high-frequency details. For \emph{global edits}, where changing the overall appearance of the image is desired, we apply the full steering coefficient uniformly at all timesteps.

For image-to-image backbones, the input image already provides a structural anchor during denoising. As a result, we do not use any timestep-dependent scheduling and instead apply the full steering coefficient throughout inference. In this setting, the distinction between local and global edits mainly affects the semantic strength of the intervention rather than the scheduling strategy: local edits modify a restricted attribute while preserving most of the source image, whereas global edits alter broader image characteristics such as overall style, illumination, or scene-wide appearance.

There is also a directional difference between the two settings. In text-to-image, we start from the original generation and add the steering vector to move the sample toward the target attribute. In image-to-image, we first obtain the edited image and then apply the negative steering direction to suppress the edit and recover a controllable range back toward the original image.
\section*{Hyperparameters Used in Runtime Calculation}

For the runtime results reported in \cref{tab:runtime_breakdown} in the main submission, we used a practical implementation setup optimized for fast inference. In the token pooling stage, we use a batch size of 100, which fits within a single H100 GPU in our setup. This stage is performed once per concept and then cached, so its cost does not need to be paid again for every new prompt. Similarly, dataset generation is also performed once per concept and cached.

For the elastic search stage, we use the following hyperparameter configuration for local edit:
\\ \texttt{height=512},
\\ \texttt{width=512},
\\ \texttt{target\_gap=0.25},
\\ \texttt{maximum\_number\_of\_iterations=25},
\\ \texttt{expand\_threshold=0.1},
\\ \texttt{maximum\_number\_of\_points=10},
\\ \texttt{starting\_number\_of\_points=4},
\\ \texttt{lam=1.0},
\\ \texttt{epsilon=0.01},
\\ \texttt{minimum\_similarity=0.15},
\\ \texttt{max\_similarity=0.40},
\\ \texttt{move\_fraction=0.5},
\\ \texttt{minimum\_value=0.0},
\\ \texttt{maximum\_value=100.0},
\\ \texttt{inference\_batch\_size=20}, and
\\ \texttt{embed\_batch\_size=64}.

For the elastic search stage, we use the following hyperparameter configuration for global/stylization edit:
\\ \texttt{height=512},
\\ \texttt{width=512},
\\ \texttt{target\_gap=0.25},
\\ \texttt{maximum\_number\_of\_iterations=25},
\\ \texttt{expand\_threshold=0.1},
\\ \texttt{maximum\_number\_of\_points=10},
\\ \texttt{starting\_number\_of\_points=4},
\\ \texttt{lam=1.0},
\\ \texttt{epsilon=0.01},
\\ \texttt{minimum\_similarity=0.25},
\\ \texttt{max\_similarity=0.40},
\\ \texttt{move\_fraction=0.5},
\\ \texttt{minimum\_value=0.0},
\\ \texttt{maximum\_value=100.0},
\\ \texttt{inference\_batch\_size=20}, and
\\ \texttt{embed\_batch\_size=64}.
\\
These hyperparameters are used for the per-prompt elastic search procedure that identifies the valid continuous steering range. We do not introduce additional nontrivial hyperparameters for the remaining stages beyond the standard inference settings of the corresponding backbones.

To further improve runtime efficiency, we utilize SGLang \cite{sglang} for serving and FlashAttention-3 \cite{flashattention3} for accelerated attention computation. These optimizations reduce overhead during generation and representation extraction, and are used throughout the reported runtime measurements.

\section*{Text-to-Image and Image-to-Image Comparison}
A practical difference between text-to-image and image-to-image backbones is whether the model already contains an explicit mechanism for preserving the input image structure during the denoising process. In purely text-conditioned generation models such as Wan~\cite{wan}, CogVideoX~\cite{cogvideox}, and FLUX~\cite{flux}, there is no internal image-conditioning that enforces structural alignment with a reference image. As a result, directly applying a large steering coefficient uniformly across all timesteps can cause the steering signal to dominate early denoising stages and alter the global scene layout. To mitigate this, we use a simple linear timestep schedule in which the effective steering strength at timestep $t$ is scaled as $\frac{t}{T}\omega$, where $T$ is the total number of timesteps and $\omega$ is the original steering coefficient. This design keeps the steering weak during the earliest denoising stages, allowing the model to first establish the coarse low-frequency structure of the scene, and gradually increases the intervention at later steps where high-frequency details and local attributes are refined. In practice, this is particularly helpful for localized edits, where preserving the original composition is essential. For more global edits, however, we do not use this progressive schedule and instead apply the original steering scale uniformly across all timesteps, since in such cases modifying the full generation trajectory is desirable.

For image-conditioned editing models \cite{qwen_image}, the situation is different. Since these models already receive an input image and are explicitly designed to preserve its structure while performing the requested modification, an additional timestep-based steering schedule is generally unnecessary. The image-conditioning mechanism already acts as a strong structural anchor throughout denoising. Therefore, for image-to-image pipelines we apply the full steering coefficient at all timesteps, without any progressive scaling.

There is also a conceptual difference in how the slider range is applied for the two settings. In text-to-image generation, we begin from the original prompt-conditioned generation and add the steering direction to move the sample away from the original image along the desired semantic axis. That is, increasing the steering coefficient progressively strengthens the target edit. In contrast, for image-to-image editing we first apply the intended edit using the editing model, and then use the negative steering direction, i.e., $-1 \times \omega$, to suppress or roll back the applied edit. This allows us to recover a continuous range between the fully edited result and a version closer to the original image. Hence, while text-to-image steering is used to diverge from the original generation toward the target attribute, image-to-image steering is used in the reverse direction to suppress the already-applied edit and thereby identify the valid slider range.

\section*{Dreamsim Scores of Quantitative Comparison}
The quantitative dreamsim scores of the compared methods is provided in \cref{tab:dreamsim}
\begin{table}
  \centering
  \setlength{\tabcolsep}{2.5pt}
  \renewcommand{\arraystretch}{1.12}
  \resizebox{\linewidth}{!}{%
  \begin{tabular}{l ccccc cccc}
    \toprule
    \textbf{Metric} &
    \multicolumn{4}{c}{\textbf{Training-based}} &
    \multicolumn{5}{c}{\textbf{Training-free}} \\
    \cmidrule(lr){2-5}\cmidrule(lr){6-10}
    &
    \textbf{SAEdit} & \textbf{Flux Slider} & \textbf{SliderEdit} & \textbf{Kontinuous Kontext} & \textbf{FluxSpace} &
    \textbf{Ours (First Token)} & \textbf{Ours (All Tokens)} & \textbf{Ours (Flux.Dev)} & \textbf{Ours (Qwen)} \\
    \midrule
    DreamSim$\downarrow$   & 0.08   & 0.23 & 0.31   & 0.21   & 0.09   & 0.06   & 0.34 & 0.20      & 0.21   \\
    \bottomrule
  \end{tabular}%
  }
  \caption{\textbf{Quantitative dreamsim comparison of editing methods.}
We report dreamsim distance of compared methods where lower score means closer perceptual similarity to the original image hence a more content preserving edit.  Methods are grouped into \textit{training-based} and \textit{training-free} settings.}
  \label{tab:dreamsim}
\end{table}

\section*{Per-Category Quantitative Comparison}

The per-category results of the quantitative comparison are reported in \cref{tab:group}.

\begin{table}[t]
\centering
\setlength{\tabcolsep}{3pt}
\renewcommand{\arraystretch}{1.12}

\resizebox{\linewidth}{!}{%
\begin{tabular}{l l c c c c c c c}
\toprule
\textbf{Edit Type} & \textbf{Metric} &
\multicolumn{4}{c}{\textbf{Training-based}} &
\multicolumn{3}{c}{\textbf{Training-free}} \\
\cmidrule(lr){3-6}\cmidrule(lr){7-9}

& &
\textbf{SAEdit} & \textbf{Flux Slider} & \textbf{SliderEdit} & \textbf{Kontinuous Kontext} &
\textbf{FluxSpace} & \textbf{Ours (Flux.Dev)} & \textbf{Ours (Qwen)} \\
\midrule

\multirow{2}{*}{\textbf{Stylization}}
& $\Delta$VQA$\uparrow$        & 0.06 & 0.28 & 0.77 & 0.53 & 0.23 & 0.34 & 0.63 \\
& $\Delta$DreamSim$\downarrow$ & 0.06 & 0.20 & 0.33 & 0.25 & 0.19 & 0.14 & 0.24 \\

\midrule

\multirow{2}{*}{\textbf{Global Edits}}
& $\Delta$VQA$\uparrow$        & 0.22 & 0.36 & 0.80 & 0.59 & 0.06 & 0.60 & 0.68 \\
& $\Delta$DreamSim$\downarrow$ & 0.09 & 0.23 & 0.34 & 0.25 & 0.04 & 0.32 & 0.24 \\

\midrule

\multirow{2}{*}{\textbf{Local Edits}}
& $\Delta$VQA$\uparrow$        & 0.32 & 0.44 & 0.68 & 0.35 & 0.22 & 0.43 & 0.61 \\
& $\Delta$DreamSim$\downarrow$ & 0.09 & 0.26 & 0.26 & 0.13 & 0.07 & 0.16 & 0.15 \\

\bottomrule
\end{tabular}%
}

\caption{\textbf{Per-category quantitative comparison.}
We report $\Delta$VQA$\uparrow$, measuring edit compliance, and $\Delta$DreamSim$\downarrow$, measuring perceptual deviation from the input image (lower indicates better identity preservation). Results are grouped by edit category: \textit{stylization}, \textit{global edits}, and \textit{local edits}.}
\label{tab:group}

\end{table}

\end{document}